\documentclass[conference]{IEEEtran}
\IEEEoverridecommandlockouts
\usepackage{cite}
\usepackage{url}
\usepackage{amsmath,amssymb,amsfonts}
\usepackage{algorithmic}
\usepackage{graphicx}
\usepackage{textcomp}
\usepackage{xcolor}
\def\BibTeX{{\rm B\kern-.05em{\sc i\kern-.025em b}\kern-.08em
    T\kern-.1667em\lower.7ex\hbox{E}\kern-.125emX}}
\begin{document}
\bstctlcite{IEEEexample:BSTcontrol}

% \title{Handling and Associating Uncertainty and Concept Drift in Streaming Scenarios for Industrial AI\\
\title{On the Connection between Concept Drift and Uncertainty in Industrial Artificial Intelligence
%\title{The Relation between Concept Drift and Uncertainty Estimation for Regression\\
\thanks{This research has received funding from the European Union’s Horizon 2020 research and innovation programme under grant agreement No: 101000162 (PIACERE project).}
}

%\author{\IEEEauthorblockN{Anonymous Authors}}

\author{\IEEEauthorblockN{Jesus L. Lobo\IEEEauthorrefmark{1}, Ibai Laña\IEEEauthorrefmark{1}, Eneko Osaba\IEEEauthorrefmark{1} and Javier Del Ser\IEEEauthorrefmark{1}$^,$\IEEEauthorrefmark{2}
}
\IEEEauthorblockA{\IEEEauthorrefmark{1}: TECNALIA, Basque Research and Technology Alliance (BRTA), 48160 Derio, Bizkaia, Spain\\
\IEEEauthorrefmark{2}: University of the Basque Country (UPV/EHU), 48013 Bilbao, Bizkaia, Spain\\
Email: jesus.lopez@tecnalia.com}
}

\maketitle

\begin{abstract}
AI-based digital twins are at the leading edge of the Industry 4.0 revolution, which are technologically empowered by the Internet of Things and real-time data analysis. Information collected from industrial assets is produced in a continuous fashion, yielding data streams that must be processed under stringent timing constraints. Such data streams are usually subject to non-stationary phenomena, causing that the data distribution of the streams may change, and thus the knowledge captured by models used for data analysis may become obsolete (leading to the so-called \emph{concept drift} effect). The early detection of the change (\emph{drift}) is crucial for updating the model's knowledge, which is challenging especially in scenarios where the ground truth associated to the stream data is not readily available. Among many other techniques, the estimation of the model's confidence has been timidly suggested in a few studies as a criterion for detecting drifts in unsupervised settings. The goal of this manuscript is to confirm and expose solidly the connection between the model's confidence in its output and the presence of a concept drift, showcasing it experimentally and advocating for a major consideration of uncertainty estimation in comparative studies to be reported in the future.
\end{abstract}


\section{Online Learning in Industrial AI}

IIoT (\emph{Industrial Internet of Things}) is considered an expanding field that refers to industrial machines and devices that are hyper-connected and that have high-performing Edge Computing functionalities. In this context, AI-based digital twins emerge from the necessity of representing a physical process or system as a digital actionable program, capable of acting as its real counterpart for practical purposes (e.g. simulation, integration, testing, monitoring, or maintenance). Digital twins generate a wealth of diverse industrial Big Data in a real-time and continuous fashion, giving rise to data streams, which are typically high-dimensional, non-stationary, and infinitely flowing. Such properties imprint particular constraints in the way algorithms can process, analyze and learn from such data streams, including their capability to continuously update their knowledge as new data arrive (\textit{Online Learning} \cite{hoi2021online}, OL). In many industrial scenarios, OL algorithms must also deal with the non-stationary nature of the distribution underneath the analyzed data streams, which may change over time as an effect of unexpected events that are exogenous to the data themselves (e.g. out of calibration IIoT sensors or a new heat source close to the industrial asset being monitored). This results in the so-called \textit{Concept Drift} \cite{sato2021survey} (CD) phenomenon, where an accurate and early detection of drifts becomes crucial. This short abstract aims to highlight the potential of Uncertainty Estimation (UE) techniques to successfully achieve this goal.

\section{The Concept Drift Phenomenon}

Formally, CD between $t$ and $t+1$ can be defined as: $P_{t} (X,y) \neq P_{t+1} (X,y)$, where $P_{t} (X, y)$ denotes the joint probability distribution at time $t$ between the input features $X$ and the target variable $y$. The joint probability can be decomposed as $P_{t} (X,y) \doteq P_{t} (X) \cdot P_{t} (y|X)$, thus changes in a data stream can be characterized by changes in $P_{t} (X)$ (\textit{feature drift}, FD) and/or $P_{t} (y|X)$ (\textit{real drift}, RD). When it occurs, a CD can provoke a degradation in the performance of models learned from drifting data streams over time. As mitigation measure, OL algorithms must be able to early detect and adapt to CD so as to promptly recover their performance. Consequently, an early CD detection becomes a crucial skill for OL algorithms in non-stationary industrial scenarios. However, detection mechanisms for the purpose differ depending on the type of CD (FD and/or RD). 

While in FD changes must be detected by using only feature information (useful in unsupervised scenarios), detection mechanisms for RD rely on the changes in the relationship between features and target (as in supervised scenarios). Unlike FD, where a change may provoke (\textit{virtual drift}) or not a drop in the model's performance, RD will always cause a model degradation. Unfortunately, labeled scenarios usually make the impractical assumption that ground truth is immediately available in the next time step after a prediction is issued by the model (test-then-train assumption). Due to this limiting assumption, digital twins and other industrial applications adopting data-based learning pipelines are in need for effective methods that can detect RD in an unsupervised manner. Among other choices, UE can be an essential approach for this purpose.

\section{Uncertainty Estimation}

UE refers to the process of quantifying the level of confidence of a model in its predicted outcome. In regression tasks, the so-called prediction interval characterizes the numerical range over which the output of a model can be distributed: a wider range implies a higher uncertainty and a lower confidence of the model in the prediction, while a narrower area reveals a higher confidence in the results. For an \textit{h}-step prediction task and a target variable \textit{y} with an estimate of the standard deviation $\widehat{\sigma}_{h}$ (as an interval estimator), the prediction interval at query time $t$ can be calculated as $\widehat{y}_{t+h \mid t} \pm c\widehat{\sigma}_{h}$, where $c$ is a parameter that depends on the coverage probabilities (e.g., $c=1.96$ for a $95\%$ prediction interval). An UE technique aims to produce a reliable estimate of the model's confidence, in the form of standard deviations, upper and lower bounds or any other numerical indicator of the support over which its output can be distributed. 

One way of generating such estimates is through quantile regression, which is especially useful when the focus is on ranking observations correctly. Some prediction algorithms can accommodate quantile formulations of their loss functions which, by means of a parameter $\alpha$, can indicate which quantile the model should be targeting during its training process. When $\alpha=0.5$, the model targets at predicting the median value of the target variable. By predicting the lower (LQ) and upper (UQ) quantiles of the target variable for a given value of $\alpha$, we are able to produce a confidence region in which the predictions are likely to reside. As such, predicting the e.g., $5$-th, $50$-th, and $95$-th quantiles requires training 3 separate models learned incrementally from the stream flows under consideration.

To date, very scarce studies have elaborated on the detection of CD by leveraging UE techniques \cite{yu2018request,baier2021detecting,wang2022concept}. Moreover, a few software frameworks consider UE to carry out an unsupervised CD detection (such as Nannyml\footnote{\url{https://www.nannyml.com/}} or Cloudera Fast Forward Labs\footnote{\url{https://www.cloudera.com/products/fast-forward-labs-research.html}}). Although the connection between an increase of the uncertainty and the occurrence of a CD can be intuitively expected, all those studies and frameworks implicitly depart from the understanding that this connection indeed exists. To the best of our knowledge, no prior research has evinced that this relationship holds.

\section{On the Relationship between CD and UE}

To expose the above relationship, a simple case study comprising a synthetic dataset has been devised to simulate an online regression task in the presence of abrupt CD. This dataset is composed of $1,000$ samples (one sample per time step) with one independent (explanatory) variable $X_t$ that follows a normal distribution, and one dependent (target) variable $y_t$ created as $Y_t = \beta_0 + \beta_1 X_t + \epsilon$, where $\beta_0$ is the intercept, $\beta_1$ is the slope of the linear relationship, and $\epsilon$ is a residual error term. A CD is induced in the dataset by simply changing the values of $\beta_0$ and $\beta_1$ at $t=500$ on, yielding two different data distributions (\emph{concepts}) and forcing the need for adapting the values $\beta_0$ and $\beta_1$ learned by two linear quantile regressors for lower ($LQ$) and upper quartiles ($UQ$) over time. Scripts are publicly available\footnote{\url{https://github.com/TxusLopez/CAI2023}}.

\begin{figure}[h!]
\vspace{-4mm}
\centerline{\includegraphics[width=0.9\columnwidth]{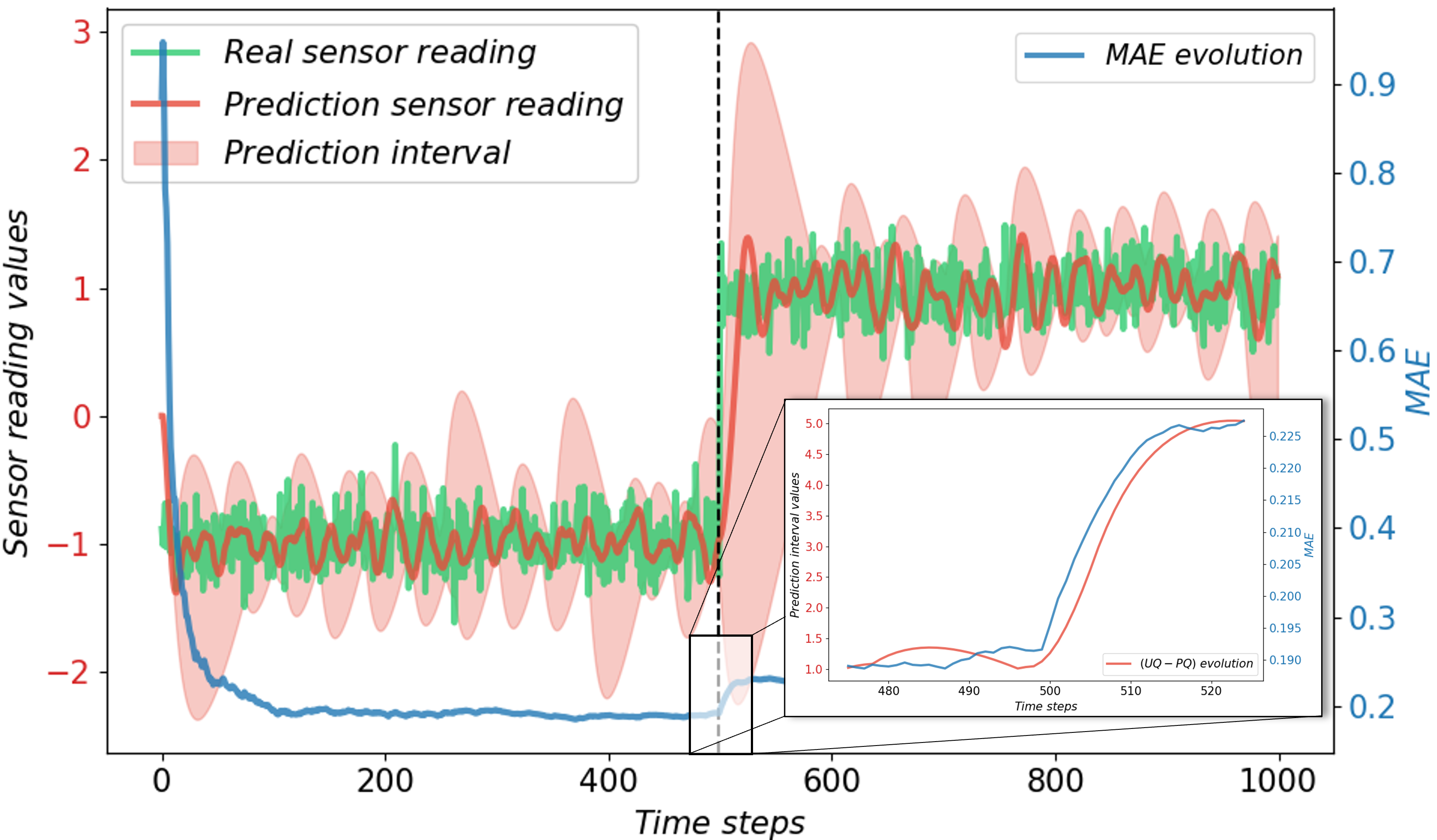}}
\vspace{-2mm}
\caption{Impact of the CD (vertical dotted line) on 1) the error metric (Mean Absolute Error, MAE), and 2) prediction interval (shaded area). In the nested plot, evolution of MAE and $(UQ-LQ)$ in the proximity of the CD.}
\vspace{-4mm}
\label{fig_SD}
\end{figure}

Figure \ref{fig_SD} verifies that there exists an increase of 1) the prediction interval and 2) the prediction error at the same time right after the drift occurs at $t=500$. When observing the nested plot in this figure (drifting period), we are able to draw the same conclusion from the high Pearson correlation coefficient (0.98) between the error metric of the model (Mean Absolute Error) and the prediction interval ($UQ-LQ$) estimated by quantile regression. This confirms the high statistical bond between the CD appearance and the model's confidence.

\section{Conclusions and Future Work}

Concept drift is a relevant problem in industrial Artificial Intelligence, with emphasis on digital twins where learning models have to operate in a real-time fashion. By developing efficient drift detectors for unsupervised data streams, the performance degradation of models resulting from varying data distributions can be counteracted. In doing so, uncertainty estimation can be a promising tool. This study has confirmed that the model's confidence can be an unsupervised indicator of the existence of a drift in regression tasks formulated over data streams, upon which effective model adaptation strategies can be triggered. In the light of these results, unsupervised drift detection mechanisms can be devised for more complex industrial scenarios subject to unexpected circumstances affecting the distribution of the modeled data, including the \textit{Open-World Learning} paradigm in which unknown data instances (such as that resulting from a FD) have to be characterized, consolidated and incorporated to the model's knowledge.

\bibliographystyle{IEEEtran}
\bibliography{CAI_2023_sub}

% Generated by IEEEtran.bst, version: 1.14 (2015/08/26)
\begin{thebibliography}{1}
\providecommand{\url}[1]{#1}
\csname url@samestyle\endcsname
\providecommand{\newblock}{\relax}
\providecommand{\bibinfo}[2]{#2}
\providecommand{\BIBentrySTDinterwordspacing}{\spaceskip=0pt\relax}
\providecommand{\BIBentryALTinterwordstretchfactor}{4}
\providecommand{\BIBentryALTinterwordspacing}{\spaceskip=\fontdimen2\font plus
\BIBentryALTinterwordstretchfactor\fontdimen3\font minus
  \fontdimen4\font\relax}
\providecommand{\BIBforeignlanguage}[2]{{%
\expandafter\ifx\csname l@#1\endcsname\relax
\typeout{** WARNING: IEEEtran.bst: No hyphenation pattern has been}%
\typeout{** loaded for the language `#1'. Using the pattern for}%
\typeout{** the default language instead.}%
\else
\language=\csname l@#1\endcsname
\fi
#2}}
\providecommand{\BIBdecl}{\relax}
\BIBdecl

\bibitem{hoi2021online}
S.~C. Hoi, D.~Sahoo \emph{et~al.}, ``Online learning: A comprehensive survey,''
  \emph{Neurocomputing}, vol. 459, pp. 249--289, 2021.

\bibitem{sato2021survey}
D.~M.~V. Sato, S.~C. De~Freitas \emph{et~al.}, ``A survey on concept drift in
  process mining,'' \emph{ACM Computing Surveys}, vol.~54, no.~9, pp. 1--38,
  2021.

\bibitem{yu2018request}
S.~Yu, X.~Wang \emph{et~al.}, ``Request-and-reverify: Hierarchical hypothesis
  testing for concept drift detection with expensive labels,'' \emph{arXiv
  preprint arXiv:1806.10131}, 2018.

\bibitem{baier2021detecting}
L.~Baier, T.~Schl{\"o}r \emph{et~al.}, ``Detecting concept drift with neural
  network model uncertainty,'' \emph{arXiv preprint arXiv:2107.01873}, 2021.

\bibitem{wang2022concept}
P.~Wang, W.~Woo \emph{et~al.}, ``Concept drift detection by tracking weighted
  prediction confidence of incremental learning,'' in \emph{2022 4th
  International Conference on Image, Video and Signal Processing}, 2022, pp.
  218--223.

\end{thebibliography}

\vspace{12pt}

\end{document}